\documentclass{article}

\usepackage{PRIMEarxiv}

\usepackage[utf8]{inputenc} 
\usepackage[T1]{fontenc}    
\usepackage{hyperref}       
\usepackage{url}            
\usepackage{booktabs}       
\usepackage{amsfonts}       
\usepackage{nicefrac}       
\usepackage{microtype}      
\usepackage{lipsum}
\usepackage{fancyhdr}       
\usepackage{graphicx}       
\graphicspath{{media/image}}     

 \title{Survey on Abstractive Text Summarization: Dataset, Models, and Metrics }
 
 \author{
     Gospel Ozioma Nnadi \href{https://orcid.org/0009-0004-4651-9876}{\includegraphics[scale=0.08]{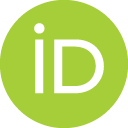}} \\
   Affiliation \\
   University of Verona \\
   Verona, Italy.\\
   \texttt{gospelozioma.nnadi@univr.it} \\
    \And
    Favio Bertini \href{https://orcid.org/0000-0001-6925-5712}{\includegraphics[scale=0.08]{ORCID.png}} \\
   Affiliation \\
   University of Parma \\
   Parma, Italy.\\
   \texttt{flavio.bertini@unipr.it} \\
 }

\begin{document}
\maketitle

\begin{abstract}
	The advancements in deep learning, particularly the introduction of transformers, have been pivotal in enhancing various natural language processing (NLP) tasks. These include text-to-text applications such as machine translation, text classification, and text summarization, as well as data-to-text tasks like response generation and image-to-text tasks such as captioning. Transformer models are distinguished by their attention mechanisms, pretraining on general knowledge, and fine-tuning for downstream tasks. This has led to significant improvements, particularly in abstractive summarization, where sections of a source document are paraphrased to produce summaries that closely resemble human expression.
  The effectiveness of these models is assessed using diverse metrics, encompassing techniques like semantic overlap and factual correctness. This survey examines the state of the art in text summarization models, with a specific focus on the abstractive summarization approach. It reviews various datasets and evaluation metrics used to measure model performance. Additionally, it includes the results of test cases using abstractive summarization models to underscore the advantages and limitations of contemporary transformer-based models. The source codes and the data are available at \cite{GitRepository}.
\end{abstract}

\textbf{Keyword:} Estractive, Abstractive, Approach, Document, Corpus, Pretrain, Finetune, Summarization.

\section{Introduction}
\label{introduction}
\begin{figure}[ht]
	  \centering	\includegraphics[height=6cm,width=0.99\linewidth]{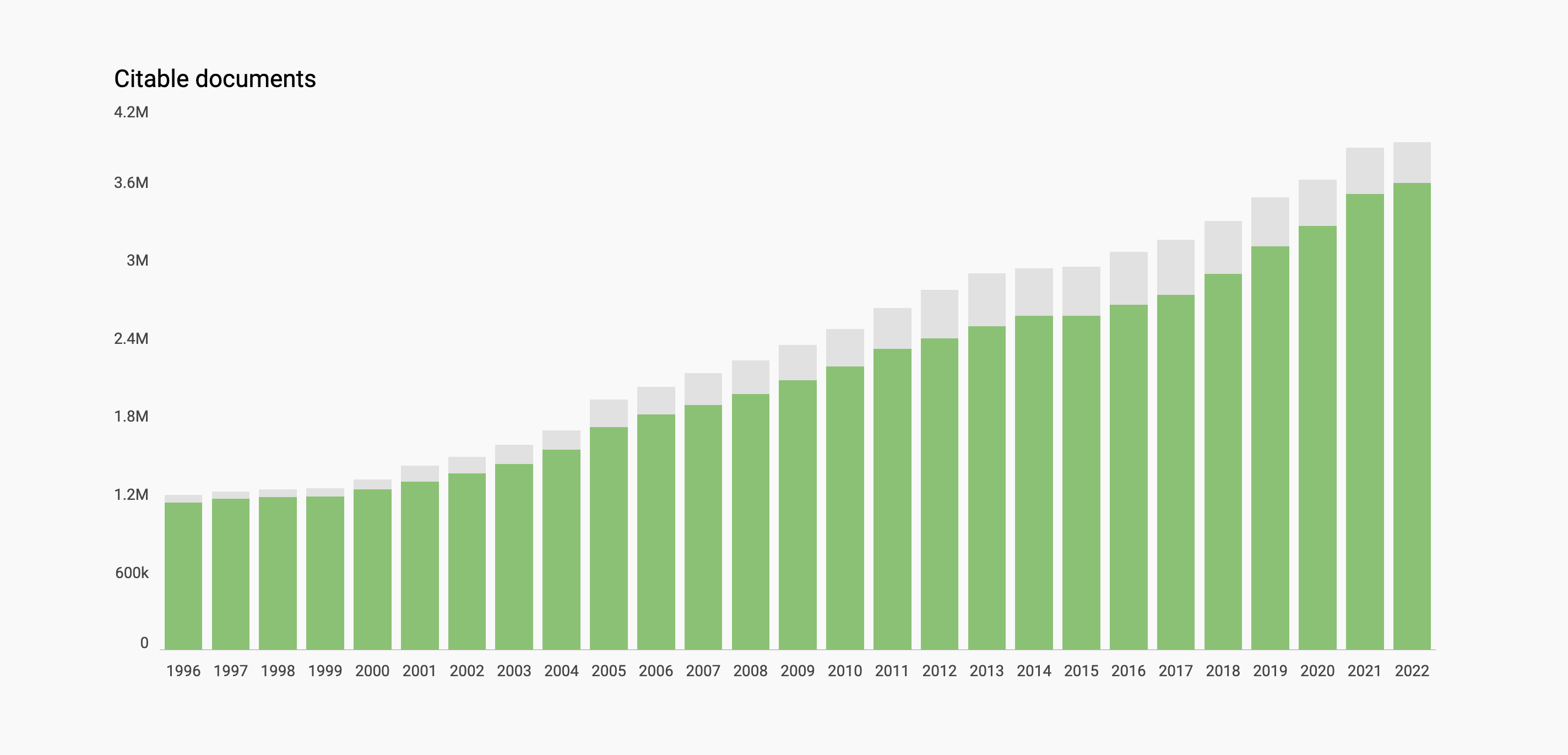}
  \caption{World report on the number of citable documents, reported by Scimagojr.}
\label{fig:Citable-doc}
  \end{figure}

\begin{figure}[ht]
	\centering	\includegraphics[height=6cm,width=0.99\linewidth]{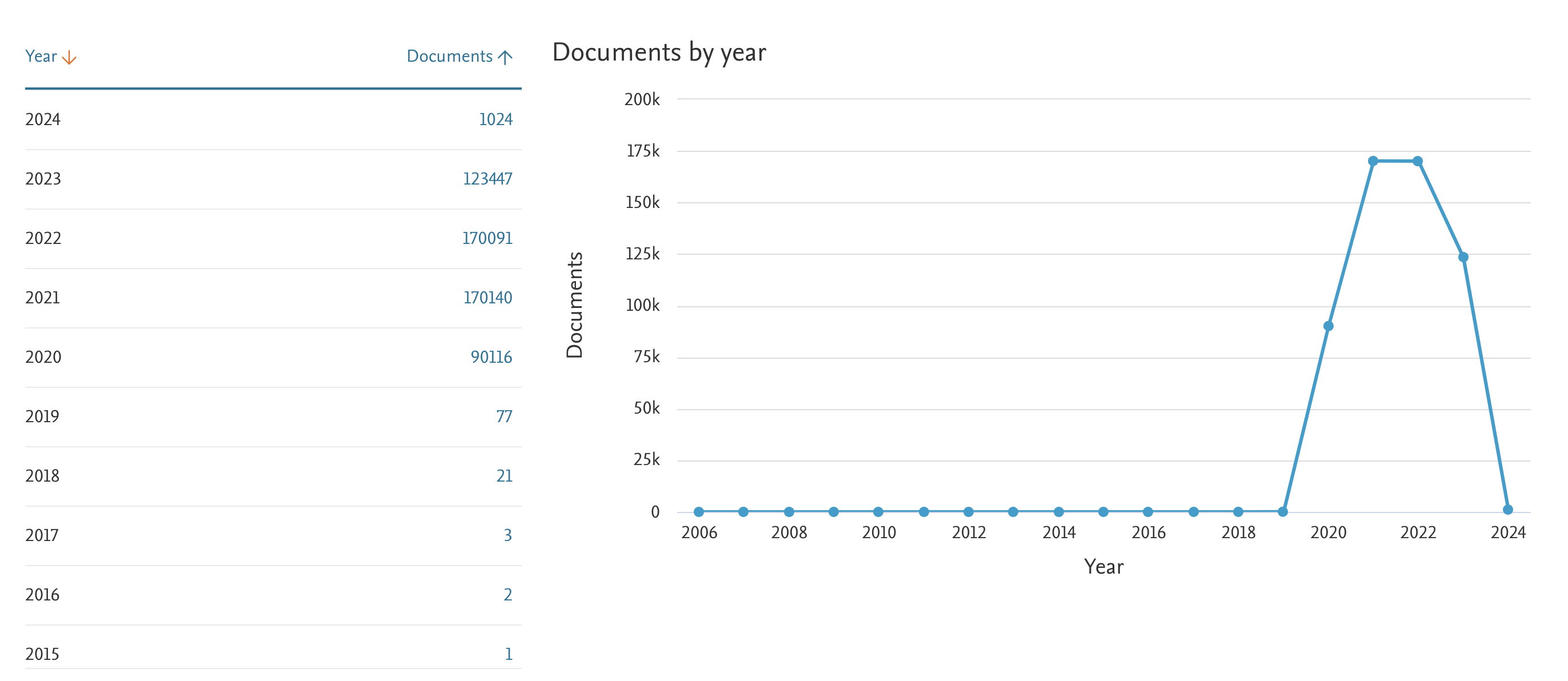}
\caption{Scopus Report of all scientific articles Covid pubblished from year 2015 to 2024.}
\label{fig:Scopus-covid}
\end{figure}

Readers and scholars often desire a concise summary (Too Long; Didn't Read - TL;DR) of texts to effectively prioritize information. However, creating document summaries is mentally taxing and time-consuming, especially considering the overwhelming volume of documents produced annually, as depicted in Figure \ref{fig:Citable-doc} by \cite{ScimagoJR:2022}, Figure \ref{fig:Scopus-covid}, \cite{Scopus:2024} reported over 100,000 scientific articles on the Corona virus pandemic in 2020, though these articles contain brief abstracts of the article, the sheer volume poses challenges for researchers and medical professionals in quickly extracting relevant knowledge on a specific topic. An automatically generated multi-document summarization could be valuable, providing readers with essential information and reducing the need to access original files unless refinement is necessary. Text summarization has garnered significant research attention, proving useful in search engines, news clustering, timeline generation, and various other applications.

\begin{figure}[ht]
		\centering	\includegraphics[width=0.99\linewidth]{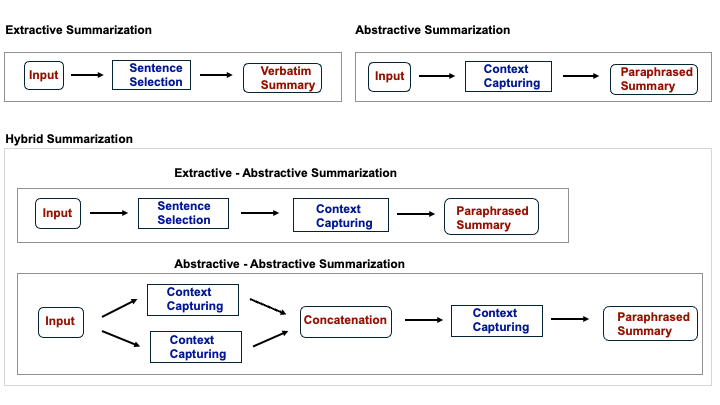}
	\caption{Text summarization approach.}
\label{fig:TsApproach}
	\end{figure}

The objective of text summarization is to create a brief, coherent, factually consistent, and readable document that retains the essential information from the source document, whether it is a single or multi-document. In \textbf{Single Document Summarization (SDS)} only one input document is used, eliminating the need for additional processing to assess relationships between inputs. This method is suitable for summarizing standalone documents such as emails, legal contracts, financial reports and so on. 
The primary goal of \textbf{Multi Document Summarization (MDS)} is to gather information from several texts addressing the same topic, often composed at different times or representing diverse perspectives. The overarching objective is to produce information reports that are both succinct and comprehensive, consolidating varied opinions from documents that explore a topic through multiple viewpoints.
 The following are examples of  multi document summarization tasks \cite{Ma:2021-Multi-document}:
\begin{itemize}
\item Query-oriented multi-document summarization involves generating a concise summary from a collection of documents that directly addresses a specific query. This process focuses on extracting information relevant to the user's question from the selected set of documents.
\item Dialogue summarization: a summary from multiple textual utterances of two or more participants, condensing a dialogue between individuals into a concise summary that captures the key points.
\item Stream summarization: summarize new documents in a continuously growing document stream, providing brief summaries of continuously incoming social media posts to capture the evolving information in real-time.
\end{itemize}
Achieving the goal of text summarization requires methods with robust capabilities to analyze input documents and identify consistent information. The methods developed for text summarization task is divided into three approach as in Figure \ref{fig:TsApproach}: 
\begin{enumerate}
\item Extractive: One of the early attempts at automatic text summarization in 1985 involved extracting important sentences based on word frequency \cite{Luhn:1958-Automatic}. This method aims to identify sentences from the source document verbatim that capture its essence.
 \item Abstractive: The abstractive approach involves crafting new sentences by paraphrasing sections of the source document, aiming to condense the text more dynamically. This is in contrast to the extractive approach, which directly pulls sentences from the source documents verbatim to capture their essence.    
 \item Hybrid: The Hybrid approach combines extractive and abstractive techniques, dividing the summarization process into two distinct phases: important content selection and then context aware paraphrasing.

\end{enumerate}

	In this article, we guide the reader through the state of the art models in text summarization, exploring the diverse datasets used for pretraining and finetuning. Additionally, we delve into various metrics employed to evaluate the quality of summaries. The article includes test cases for both single-document and multi-document scenarios, with the aim of showcasing the advantages and the limitations of abstractive text summarization models, particularly for pre-trained transformer-based models.

    \section{Summarization Tasks} \label{State of the art}

    The current leading text summarization models stand out based on their approach and the nature of the source document (single or multi-document). Moreover, these models exhibit distinctions rooted in the self-attention quadratic effect found in existing abstractive summarization models. Further granularity is achieved by categorizing them based on the dimension of the document, distinguishing between short and long documents.
    \subsection{Short Document Summarization vs Long Document Summarization}
    The length of a document is a crucial factor in text summarization models. While sequence-to-sequence models based on Recurrent Neural Networks (RNNs) can sequentially read all tokens and store context, the attentions may become dispersed over long sequences, potentially degrading performance. Transformer-based models like BART \cite{Lewis:2019BART}, PEGASUS \cite{Zhang:2020Pegasus}, and others face a limitation with a fixed maximum input sequence (typically 512 words, extended to 1024 words) due to the computational complexity of self-attention, which grows quadratically with sequence length.
    
    In handling long documents for summarization, short document summarization models may face challenges leading to potential information loss. These challenges include:
    \begin{itemize}
        \item \textbf{Truncation}: Disregarding text exceeding a predefined threshold. Relevant information beyond the threshold is ignored, potentially leading to incomplete or skewed summaries.
        \item \textbf{Chunking errors}: Dividing the document into chunks, processing each separately, and then combining activations using a task-specific model. Contextual information in parts of the document split into chunks may not be fully considered, resulting in gaps or inconsistencies in the summary.
        \item \textbf{Cascading errors from the two stage approach}: Employing a two-stage model where the first stage retrieves relevant documents, passed on to the second stage for extraction. The quality of the summary heavily relies on the first stage, which may mistakenly select non-relevant words, leading to inaccuracies in the final summary.
    \end{itemize}
    
    In recent developments, novel global attention transformer-based models such as Longformer \cite{Beltagy:2020Longformer}, BigBird \cite{Zaheer:2021Big}, and LongT5 \cite{Guo:2022LongT5} have been introduced. These models are specifically designed and pretrained to handle long sequences. Unlike the original transformer model \cite{Vaswani:2017Attention}, which employ full self-attention and face quadratic growth in memory and computational requirements with sequence length, making them impractical for processing long documents, these new models modify the transformer architecture with an attention pattern that scales linearly with the input sequence.
    
    This modification makes these models versatile for processing long documents, allowing them to handle sequences of up to 16,000 words. To provide context, this corresponds to approximately 32 pages for single-spaced text or 64 pages for double-spaced text with a font size of 12 points. Additionally, BigBird demonstrates the capability to leverage pretraining for learning powerful contextual representations, even for Deoxyribonucleic Acid (DNA) fragments \cite{Zaheer:2021Big}.

    \subsection{Single Document Summarization vs Multi Document Summarization }
    
    Single document summarization and multi document summarization share the common goal of condensing documents into concise and coherent summaries. They utilize similar summarization construction types as illustrated in Figure \ref{fig:TsApproach}, learning strategies, evaluation indexes, and objective functions, aiming to minimize the distance between machine-generated and human-annotated summaries. Transformer-based models, such as BERT \cite{Devlin:2019Bert}, BART \cite{Lewis:2019BART}, and PEGASUS \cite{Zhang:2020Pegasus}, have improved abstractive text summarization for single documents. Early methods for multi document summarization applied single document summarization techniques however, there are significant differences between single document summarization and multi document summarization, leading to the development of modified transformer-based models  \cite{Vaswani:2017Attention} such as \textit{PRIMERA} \cite{Xiao:2022Primera}, \textit{CENTRUM} \cite{Xiao:2022Primera}, and more.
    
          There are five aspects highlighting the distinctions between single document summarization and multi document summarization:
        \begin{enumerate}
        \item More diverse input document types; multi document summarization deals with multiple input documents from various sources, introducing diversity in the types of documents being summarized.
        \item Insufficient methods in single document summarization models to capture cross-document relations; multi document summarization often employs clustering and hierarchical graph methods to capture cross-document relations, a challenge not adequately addressed in single document summarization models.
        \item High redundancy and contradiction across input documents; multiple input documents used in multi document summarization are likely to contain more contradictory, redundant, and complementary information compared to single document scenarios.
        \item Larger searching space and limited training data; multi document summarization models face a larger searching space but often lack sufficient training data compared to single document summarization. This presents obstacles for deep learning models to learn adequate representations, although progress has been made with the availability of new large-scale labeled and unlabeled data in recent years.
        \item Lack of evaluation metrics specifically designed for multi document summarization; existing single document summarization evaluation metrics may not effectively evaluate the relationship between the generated abstract and different input documents in the context of multi document summarization.
    \end{enumerate}

    Multi-document summarization tasks involve processing multiple input documents from diverse sources, typically categorized into three groups:
    \begin{itemize}
        \item Many Short Sources: This involves a large number of brief documents. For instance, summarizing product reviews entails creating a concise summary from numerous individual reviews of the same product or service.
    
        \item Few Long Sources: This category focuses on generating a summary from a small number of lengthy documents, such as synthesizing a group of news articles or creating a Wikipedia-style entry from multiple web-based articles.
    
        \item Hybrid Sources: These combine one or a few long documents with several shorter ones. Examples include summarizing news articles accompanied by reader comments or producing a scientific summary based on a lengthy paper and multiple short citations.
    \end{itemize}	
    
    In multi document summarization tasks, two common methods are used to concatenate multiple input documents:
    \begin{itemize}
            \item Flat concatenation, treats the multi document summarization task as a single document summarization task. Models processing flat concatenated documents need a strong ability to handle long sequences.
            \item Hierarchical concatenation, preserves cross-document relations, enhancing model effectiveness. Existing methods include:
            \begin{itemize}
                \item Document-level condensing in a cluster, where documents are processed separately or at the word or sentence level inside document clusters \cite{Antognini:2019Learning, Nayeem:2018Abstractive}. Models like \textit{PRIMERA} \cite{Xiao:2022Primera} and \textit{CENTRUM} \cite{Xiao:2022Primera} use document clustering.
                  \item Sentence relation graph to model the hierarchical relations among multi-documents \cite{Yasunaga:2017Graph}. Graph construction methods use sentences as vertices and edges to indicate sentence-level relations. Graph structures like cosine similarity, discourse graph, semantic graph, and heterogeneous graph serve as external knowledge to enhance multi document summarization models. Models such as \textit{REFLECT} \cite{Song:2022Improving} and BART-Long \cite{Pasunuru:2021Efficiently-Summarizing} are graph-based.
                \end{itemize}

        \end{itemize}	
                 
     \subsection{Extractive vs Abstractive vs Hybrid Summarization Approach} 
     The extractive, abstractive, and hybrid summarization approaches define the construction types (Figure \ref{fig:TsApproach}) used by both single document summarization and multi document summarization methods to handle input documents, process information, and generate the final summary. Notably, the abstractive approach has experienced substantial advancements, particularly in the last six years, owing to continuous improvements in deep learning models with the introduction of transformer \cite{Vaswani:2017Attention}.\\
    
     \textbf{Extractive Text Summarization}: Methods typically involve assigning each sentence a saliency score and ranking sentences in the document based on these scores. The scoring is often determined using a combination of statistical and linguistic features, such as term frequency, sentence position, cue words, stigma words, and topic signature. Machine learning models, including both unsupervised and supervised methods, have been employed for sentence extraction. Approaches like Maximal Marginal Relevance (MMR), Latent Semantic Analysis (LSA), and graph-ranking methods have also been used to iteratively extract key phrases and sentences.
    
     One challenge with extractive summarization is the potential for redundancy. Recent methods in multi document summarization address this issue by simplifying and generating ideograms representing the meaning of each sentence, comparing their shapes and positions, and then summarizing based on user-supplied parameters for equivalence and relevance.
     
     Extractive summarization approach differs from abstractive summarization approaches in the following way:
     \begin{itemize}
        \item it can be viewed as word or sentence-level classifiers, involving sentence ranking and selection. Recent models apply reinforcement learning \cite{Wu:2018Learning}; \cite{Dong:2018BanditSum} strategies to optimize the model directly on task-specific, non-differentiable reward functions.
        \item the models are independent of any language and do not require additional linguistic knowledge.
        \item it often produce non coherent, redundant summaries.
        \item it produce summaries that are composed of exact words copied from the source document.
    \end{itemize}	 
         
    Some of the models developed for \textbf{Single Document - Extractive Text Summarization} task are the following:
    
         \begin{itemize}
            \item NEUSUM \cite{Zhou:2018Neural} employs a joint scoring and selection strategy. It constructs a hierarchical representation of the document and incorporates the partially generated summary at each step of the sentence selection process.
            \item BanditSum \cite{Dong:2018BanditSum} approaches extractive summarization as a contextual bandit problem, where the document represents the context, and selecting a sequence of sentences for the summary constitutes the action.
            \item LATENT \cite{Zhang:2018bNeural-latent}, introduces a latent variable extractive model. It views relevance labels of sentences in a document as binary latent variables. The latent model maximizes the likelihood of human summaries given the selected sentences.
            \item REFRESH \cite{Narayan:2018Ranking-sentences}, propose using Reinforcement learning-based system (\cite{Williams:1992Simple-statistical}). It approximates the search space during training by focusing on combinations of individually high-scoring sentences. The model globally optimizes the ROUGE metric for evaluation.	
            \item RNES \cite{Wu:2018Learning}, aims to enhance the coherence of extractive summaries by introducing a coherence model and incorporating it into the reinforcement learning framework along with ROUGE-based rewards. This approach seeks to improve the overall quality and coherence of the generated summaries.
            \item JECS \cite{Xu:2019Neural-extractive}, combines the extraction of individual sentences with a constituency-based approach to compression, enabling the model to capture more global dependencies and structural information in the document. The goal is to improve the overall coherence and content selection in the generated summaries.
            \item STRASS \cite{Bouscarrat:2019STRASS}, focuses on selecting sentences with embeddings that are most semantically similar to the document embedding, and it fine-tunes a linear transformation to improve alignment with reference summaries, as measured by ROUGE-1 similarity. This approach aims to enhance the content and coherence of the extractive summary generated by the model.
            \end{itemize}

            Some of the models developed for \textbf{Multi Document - Extractive Text Summarization} task are the following:
     \begin{itemize}
     \item MTSQIGA \cite{Mojrian:2021Novel-extractive}, treats multi-document summarization as a binary optimization problem, employing a modified quantum-inspired genetic algorithm (QIGA) to find the best solution. The optimization involves a linear combination of coverage, relevance, and redundancy factors, integrating six sentence scoring measures.
     \item CDDS \cite{Alguliev:2013CDDS}, constraint-driven Document Summarization focuses on generating summaries with maximum content coverage and high diversity. It utilizes user-provided constraints and tunes constraint parameters to achieve these objectives.
     
     \item MA$\&$MR \cite{Alguliyev:2015MAMR}, employs a combination of symmetric and asymmetric similarity measures, including weighted harmonic mean of cosine similarity and overlap similarity. It formulates text summarization as a boolean programming problem, optimizing it using the differential evolution (DE) algorithm.
     
     \item MCMR \cite{Alguliyev:2011MCMR}, represents multi-document text summarization as an integer linear programming (ILP) problem, aiming to optimize both coverage and redundancy simultaneously. It employs two optimization algorithms, namely branch-and-bound ($B\&B$) and particle swarm optimization (PSO), to globally solve the problem.
     
     \item PPRSum \cite{Liu:2008Personalized-pagerank}, considers multiple features to select salient sentences in a corpus. It involves training a model based on sentence global information using Naïve Bayes, generating a relevance model based on the query of each corpus, and computing personalized prior probability based on both models. It applies PPR to select the most relevant sentences.
    
     \item LFIPP \cite{Alguliyev:2011bSentence-Selection}, formulates sentence-extraction-based summarization as a linear fraction of sentence similarities, considering both sentence-to-document and sentence-to-sentence similarities. This approach aims to select salient sentences from the input document collection while reducing redundancy in the generated summary.
    \end{itemize}

    \textbf{Abstractive Text Summarization}: The evolution of neural architectures, coupled with introduction of pre-training and transfer learning methodologies \cite{Peters:2018Deep} \cite{Devlin:2019Bert}, along with the availability of large-scale supervised datasets \cite{Nallapati:2016Abstractive-Text} \cite{Grusky:2018bNewsroom} \cite{Narayan:2018Topic-aware} \cite{Sharma:2019Entity-Driven}, has led to the dominance of deep learning-based approaches in abstractive text summarization. Leading models leverage self-attention transformer blocks in an encoder-decoder structure \cite{Liu:2019Hierarchical} \cite{Liu:2019Pretrained} \cite{Zhang:2020Pegasus}, incorporating attention and copying mechanisms \cite{See:2017Pointer-Generator} \cite{Cohan:2018Discourse-aware}, and multi-objective training strategies \cite{Guo:2018Layer-Specific}, including reinforcement learning techniques \cite{Kryscinski:2018Improving}, to capture semantics and generate comprehensive summaries.
    
    Transformer-based models like BERT \cite{Devlin:2019Bert}, BART \cite{Lewis:2019BART}, PEGASUS \cite{Zhang:2020Pegasus}, have demonstrated improved performance not only in text summarization but across various natural language processing tasks. The success of transformers is attributed to pre-training using self-supervised objectives on large text corpora, enabling transfer learning to downstream tasks. The self-attention component allows the model to capture contextual information from the entire sequence. 
    
    Abstractive text summarization differs from the extractive text summarization approach in the following way:
    \begin{itemize}
        \item it involves sentence paraphrasing, generating new sentences.
        \item it uses advanced machine models, such as encoder-decoder, auto-regressor, and reinforcement learning architectures, capture context effectively. Benefits from pre-training to transfer learning and distillation knowledge for improved performance and efficiency, and speed-up; \textbf{Transfer Learning} \cite{Raffel:2023Exploring} provides a general-purpose “knowledge” to improve performance on downstream tasks, while \textbf{Knowledge Distillation} \cite{Niccolai:2020Distillation} transfers knowledge from a large model to a smaller one without loss of validity, resulting in a more deployable and resource-efficient model.
        \item it produces potentially more fluent summaries, better at reflecting conflicting information, and addresses redundancy issues present in the extractive summarization approach.
        \item but the factuality of the summries produce is still a challenge \cite{Cao:2018Fact-Aware}; \cite{Goodrich:2019Factual-Accuracy}; \cite{Falke:2019Ranking-Generated}; \cite{Kryscinski:2019Neural-Text}; they are liable to reproduce factual details inaccurately.
         \end{itemize} 
    
        \textbf{Abstractive Text Summarization Models for Single Document - Short Document}: Some of the models developed for this task are;
    \begin{itemize}
        \item BertSum-abs \cite{Liu:2019Pretrained}, BERT (Bidirectional Encoder Representations from Transformers) \cite{Devlin:2019Bert} extends the concept of word embeddings with BERT, a model that learns contextual representations from large-scale corpora using a language modeling objective. BertSum-abs introduces a document-level encoder on top of BERT.
        \item Pointer Generator \cite{See:2017Pointer-Generator}, propose a variation of Long Short-Term Memory (LSTM) \cite{Hochreiter:1997LONG-SHORT-TERM} encoder-decoder models known as the Pointer Generator Network. This model allows the decoder to choose between generating a word from the vocabulary or copying a word from the input. It includes a coverage mechanism to prevent repetitive attention to the same part of the source document.
        \item Transformer + Pointer Generator Network \cite{Martin:2020Review-Summarization}, enhances the pointer generator network and the transformer. Techniques such as n-gram blocking and coverage loss are used to create a more accurate model that reduces repetition. The copy mechanism enables the model to copy words from source texts, reducing unknown word tokens and incorrect factual representations.
        \item Improve-abs \cite{Kryscinski:2018Improving}, extend the model of \cite{Paulus:2017Deep-Reinforced}, by incorporating an external LSTM language model into the decoder. It introduces a reinforcement learning-based objective during training.
        \item ROUGESal \cite{Pasunuru:2018Multi-Reward} introduces a salience reward based on keyphrases and an entailment-based reward, complementing the traditional ROUGE-based reward within a REINFORCE framework. These rewards are optimized simultaneously using alternating mini-batches.
        \item Multi-task (Ent + QG) \cite{Guo:2018Layer-Specific}, it employs a specialized multi-task architecture to efficiently integrate question generation and entailment generation as auxiliary tasks.
        \item Closed book decoder \cite{Jiang:2018Closed-Book}, extends the Pointer Generator Network by incorporating a copy-less and attention-less decoder during training. This modification encourages the encoder to be more selective in encoding the most salient content.
        \item T5 (Text-to-Text Transfer Transformer) \cite{Raffel:2023Exploring}, is an encoder-decoder model pre-trained with a causal language modeling objective and fine-tuned for various downstream tasks. It employs a multi-task mixture of unsupervised and supervised tasks, converting each task into a text-to-text format.
        \item GPT-2 \cite{Ziegler:2020Fine-Tuning}, is a generative pretraining model trained with a causal language modeling objective on a diverse corpus of unlabeled text, followed by discriminative fine-tuning on specific tasks.
        \item UniLM \cite{Dong:2019Unified-Language}, utilizes shared model parameters across different language modeling objectives. Various self-attention masks control access to context for each word token.
        \item BART (Bidirectional and Auto-Regressive Transformers) \cite{Lewis:2019BART}, is trained by corrupting documents using random shuffling of the original order of sentences and using a novel in-filling scheme. It optimizes a reconstruction loss between the decoder’s output and the original document.
        \item PEGASUS (Pre-training with Extracted Gap-sentences for Abstractive Summarization Sequence-to-sequence) \cite{Zhang:2020Pegasus}, is pre-trained with a self-supervised gap-sentence-generation objective. It involves masking entire sentences from the source document, concatenating them, and using the result as the target “summary,” with important sentences removed and then generated from the remaining sentences.
    \end{itemize}

    \textbf{Abstractive Text Summarization Models for Single Document - Long Document}: Several models have been developed to address the challenges of abstractive text summarization for long documents. Some notable models include:
    \begin{itemize}
         \item BigBird PEGASUS \cite{Zaheer:2021Big}, introduced a sparse attention mechanism to handle very long sequences efficiently. By reducing quadratic dependency to linear, it utilizes global tokens for long-context processing.
         \item LongT5 \cite{Guo:2022LongT5}, introduced an efficient Text-To-Text Transformer-based neural model designed for handling long sequences. It integrates pre-training strategies from PEGASUS into a scalable T5 architecture. LongT5 employs the Transient Global (TGlobal) attention mechanism, mimicking long-input transformers' local/global attention without requiring additional side-inputs.
          \item Longformer \cite{Beltagy:2020Longformer}, presents a novel sparse attention mechanism, incorporating both local and global attention, to efficiently store context. It reduces self-attention's quadratic time and memory complexity to linear by sparsifying the full self-attention matrix based on a specified attention pattern.
            \item Deep Communicating Agents (DCA) \cite{Celikyilmaz:2018Deep-Communicating}, proposed an abstractive system featuring multiple agent LSTM encoders. To encode a long text, the task is divided among collaborating agents, each responsible for a subsection. The decoder, equipped with a hierarchical attention mechanism over the agents, uses contextual agent attention to smoothly integrate information from multiple agents during decoding. The model is trained end-to-end using self-critical reinforcement learning for generating focused and coherent summaries.
        \end{itemize}
    
    \textbf{Abstractive Text Summarization Model for Multi Document}Several models have been developed to address the challenges of abstractive text summarization for multi-document scenarios. Notable models include:
    \begin{itemize}
        \item GraphSum \cite{Li:2020GraphSum}, employs graphs to encode documents, capturing cross-document relations crucial for summarizing long documents. It utilizes graphs to guide the summary generation process, enhancing coherence and conciseness.
        \item BART-Long \cite{Pasunuru:2021Efficiently-Summarizing}, incorporates graph information into the pre-trained encoder-decoder model, enabling it to scale efficiently to large input documents often encountered in summarizing news clusters. The model can process auxiliary graph representations derived from multi-document clusters.
         \item HT (Hierarchical Transformer) \cite{Liu:2019Hierarchical}, learns latent dependencies among textual units by utilizing cross-document relationships through an attention mechanism. This approach allows the sharing of information, going beyond simple concatenation and processing of text spans as a flat sequence.
          \item PRIMERA \cite{Xiao:2022Primera}, emphasizes training the model to identify and aggregate key information from a cluster of related documents during pre-training. It introduces a Gap Sentence Generation (GSG) objective called Entity Pyramid, which involves masking salient sentences across the entire cluster and training the model to generate them, thereby promoting the identification of crucial information across documents.
           \item CENTRUM \cite{Xiao:2022Primera}, proposes a distinct strategy from PRIMERA for teaching multi-document summarization models to identify and aggregate salient information across a document cluster during pre-training. It employs the centroid of the cluster as a synthetic summary to guide the model's learning process.
        \end{itemize}

    \textbf{Hybrid Text Summarization}
    
    Hybrid summarization methods offer a versatile approach to capturing the semantic complexities of documents, incorporating features commonly associated with both extractive and abstractive summarization methods. These models typically operate in two stages: extractive-abstractive and abstractive-abstractive.
    \begin{itemize}
        \item Extractive-abstractive, an extractive summarization model is employed to select salient sentences. Subsequently, an abstractive summarization model is used to generate the final summary. This hybrid approach leverages the strengths of both extractive and abstractive methods.
         \item Abstractive-abstractive, involves using an abstractive summarization model during content selection or text reduction operations. Following this, another abstractive summarization model is utilized to generate the final summary. This approach emphasizes the use of abstractive methods throughout the summarization process.
        \end{itemize}

    \textbf{Hybrid Text Summarization Models for Single Document Summarization} Some hybrid models developed for single document summarization:
    \begin{itemize}
        \item Fast-abs-rl \cite{Chen:2018Fast-Abstractive}, initially extracts salient sentences using a Pointer Network and subsequently rewrites them with a Pointer Generator Network. Along with maximum likelihood training, a ROUGE-L reward is applied to update the extractor through the REINFORCE algorithm \cite{Williams:1992Simple-statistical}.
        \item Bottom-Up \cite{Gehrmann:2018Bottom-Up}, presents a bottom-up approach in which a content selection model limits the copy attention distribution of a pre-trained Pointer Generator Network during inference. The content selector is used to identify which phrases from the source document should be included in the summary.
         \item Unified-ext-abs \cite{Hsu:2018Unified-Model}, proposes utilizing the probability output from an extractive model as sentence-level attention to adjust the word-level attention scores of an abstractive model. It introduces an inconsistency loss to promote alignment and consistency between these two levels of attention.
         \item SENECA \cite{Sharma:2019Entity-Driven}, proposes the use of an entity-aware content selection module combined with an abstractive module to produce the final summary.
    \end{itemize}

    \textbf{Hybrid Text Summarization Models for Multi Document Summarization} One of the recent models developed for multi-document summarization is:
     \begin{itemize}
          \item REFLECT \cite{Song:2022Improving-Multi-Document}, proposed an extract-then-abstract transformer framework that leverages pre-trained language models. It constructs a hierarchical extractor for salient sentence selection across documents and an abstractor for rewriting the selected contents as summaries.
         \end{itemize}

\section{Methods}
         Understanding human language in computer programs involves representing words and their contexts, a crucial aspect of semantics. The introduction of Neural Networks has enchance the representation of contextual word vectors that captures only what a word means in a context. Such that, the collection of contexts a word type is found in, provides clues about its meaning(s) \cite{Smith:2020Contextual-Word}. 
         These vectors play a vital role in various natural language processing tasks, including information retrieval, document classification, question answering, named entity recognition, parsing, and text summarization.

             \begin{description}
             \item[Word embeddings]: Word embedding methods, such as Word2Vec and GloVe, are non-contextual vectors retrieved from a lookup table. To derive contextual information of a document, hidden neural layers can be incorporated.
             \begin{itemize}
             \item \textbf{Word2Vec}:  \cite{Mikolov:2013Efficient} is trained to learn vector representations that predict the probability of surrounding words occurring given a center word (SkipGram) or vice versa (Continuous Bag of Words - CBoW). In this context, the surrounding words are often referred to as context words, as they appear in the context of the center word.
             \item \textbf{GloVe}: \cite{Pennington:2014GloVe} is trained by considering the global statistical information of word co-occurrences. Unlike Word2Vec, which focuses on local context, GloVe is designed to capture global semantic relationships between words.
           \end{itemize}
             \item[Contextual embeddings]: BERT \cite{Devlin:2019Bert} and ELMo \cite{Peters:2018Deep}, generate different vector representations for the same word in different sentences based on the surrounding words, capturing contextual meaning.		\begin{itemize}
             \item \textbf{ELMo}: Which stands for “embeddings from language models,” brought a powerful advancement in the form of word token vectors, that is, vectors for words in context, or contextual word vectors that are pre-trained on large corpora. There are two important insights behind ELMo: \\
             \begin{enumerate}
               \item	if every word token is going to have its own vector, then the vector should depend on an arbitrarily long context of nearby words. \\
                   \item	it trains one neural network for left contexts (going back to the beginning of the sentence a token appears in) and another neural network for right contexts (up to the end of the sentence). Longer contexts, beyond sentence boundaries, are in principle possible as well.
             \end{enumerate}
              \item \textbf{BERT}: Bidirectional Encoder Representations from Transformers, a successor to ELMo, introduced innovations and achieved significant error reduction. BERT learns from bidirectional contexts, improving contextual understanding of words.
             The embeddings are generated by training models using language modeling objectives such as masked word prediction and Gap Sentence Generation. They produce embeddings at different layers, with intermediate layers offering more effective representations for semantic tasks.	
             \end{itemize}
             \end{description}
         
         \subsection{Transformer-based Text Summarization Models}\label{Transformer-based}
         Neural networks leveraging contextual embeddings, such as BERT \cite{Devlin:2019Bert}, BART \cite{Lewis:2019BART}, and PEGASUS \cite{Zhang:2020Pegasus}, are capable of capturing word context and generating summaries for a given text. In contrast, Convolutional Neural Network (CNN) based models \cite{Oshea:2015IntroductionCNN} are less effective at processing sequential data compared to Recurrent Neural Network (RNN) based models \cite{Schuster:1997Bidirectional}. However, RNN models face limitations in parallel computing because they heavily rely on results from previous steps. 
         Additionally, RNNs struggle with processing long sequences, as earlier information tends to fade over time. Transformer-based architectures \cite{Vaswani:2017Attention} provide a solution to these issues. The self-attention mechanism of transformers, along with the flexible attention mechanisms (local and global attention) introduced in \cite{Beltagy:2020Longformer}, offer natural advantages for parallelization and effectively retain long-range dependencies. 
         These transformer-based models have demonstrated promising results in text summarization tasks.
         \textbf{Transformer} \cite{Vaswani:2017Attention} encoder-decoder architecture:
         \begin{itemize}
           \item Input is a positional encoded word embedding and the final decoder output is ran through a linear followed by a softmax layer to convert the outputs to probabilities.
           \item Positional encoding keeps track of the order of the input sequence. 
           \item Encoder: N identical layers. Each layer has two sub-layers a multi-head self-attention mechanism, and a simple, position wise fully connected feed-forward network.
           \item Decoder: Consists of N identical layers, with an additional third sub-layer that applies multi-head attention to the output generated by the encoder stack.
           \item Masked Multi-Head Attention: masks future positions in the output preventing the decoder from peeking into the future during training.
         \end{itemize}
         To address the quadratic memory growth associated with self-attention when handling document-scale sequences, Longformer's \cite{Beltagy:2020Longformer} local and global attention mechanisms allow to scale memory usage linearly. This enables their application in text summarization tasks for long document sequences or multi-document summarization \cite{Pasunuru:2021Efficiently-Summarizing}.
         Tranformer based model with 6 encoder and 6 decoder layers is often referred to as a base model while a large model is a model with 12 or more encoder and decoder layers.
         \textbf{Pre-trained language models (LMs)}: The transformer model allows for pretraining on large text corpora, the pretrained model have shown great successes when finetuned for downstream natural language processing tasks including text summarization.  
           The following are some of the self supervised objective function for pre-training transformers based models \cite{HuggingFace}:
             \begin{itemize}
             \item \textbf{Masked Language Modeling (MLM)}: Encoder input tokens are randomly replaced by a mask tokens and have to be predicted by the encoder. BERT \cite{Devlin:2019Bert} use this self supervised objective to randomly mask and predict masked token.
             \item \textbf{Gap Sentence Generation (GSG)}: Whole encoder input sentences are replaced by a second mask token and fed to the decoder, but which has a causal mask to hide the future words like a regular auto-regressive transformer decoder. PEGASUS \cite{Zhang:2020Pegasus} use this self supervised objective, important sentences are masked and are generated together as one output sequence from the remaining sentences.
             \item \textbf{Causal language modeling (CLM)}: Which is the traditional auto-regressive training. One of the languages is selected for each training sample, and the model input is a sentence of 256 tokens, that may span over several documents in one of those languages. T5 \cite{Raffel:2023Exploring} introduced this self supervised objective.
             \item \textbf{Combination of MLM and translation language modeling (TLM)}: This consists of concatenating a sentence in two different languages, with random masking. To predict one of the masked tokens, the model can use both, the surrounding context in language 1 and the context given by language 2.
             \end{itemize}

\section{Metrics}
Robust and unbiased evaluation methods are beneficial to many Natural Language Processing task such as, text-to-text (machine translation and summarization), data-to-text (response generation), and image-to-text (captioning)  \cite{Gatt:2018Survey}. Given that, the goal of text summarization models is to automatically generate succinct, fluent, relevant, and factual consistent summaries, a generated summary can be evaluated based on quality dimensions or perspectives, that are grouped into following categories; semantic overlap (informativeness, semantic coverage, and relevance), linguistic quality (fluency, coherence), adequacy and factual correctness (factuality):
\begin{itemize}
	\item \textbf{Informativeness (INFO)}: How well the generated hypothesis captures the key ideas of the source text. 
	\item \textbf{Relevance (REL)}: How consistent the generated hypothesis is with respect to the source text.
	\item \textbf{Fluency (FLU)}: Refers to whether the text is free from formatting issues, capitalization errors, or glaring grammatical mistakes (e.g., fragments, missing components) that could hinder readability.
	\item \textbf{Coherence (COH)}: Refers to the extent to which the text logically progresses from sentence to sentence, forming a coherent and cohesive body of information on a particular topic.
	\item \textbf{Factuality (FAC) / Consistency}: Whether the generated hypothesis contains only statements entailed by the source text. 
	\item \textbf{Semantic Coverage (COV)}: How many semantic content units from reference texts are covered by the generated hypothesis. 
	\item \textbf{Adequacy (ADE)}: Whether the output conveys the same meaning as the input sentence, and none of the message is lost, added, or distorted.
	\end{itemize}

For text summarization, the gold-standard method for evaluating summary is human evaluation; human annotators assess the generated texts’ quality. Human evaluations of these task are extensive but expensive and can take long time to finish, involves human labor that can not be reused. Therefore human evaluation of summaries can be quite infeasible, to reduce the cost of human evaluation without introducing bias, evaluating the quality of automatically generated summaries is required to be done automatically with automatic evaluation methods that is quick, inexpensive, language-independent, correlates highly with human evaluation, and has little marginal cost per run.

	The following automatic metrics were introduced for single document summary. Metrics, such as S3, SummaQA, SMS, CHRF, and METEOR tended to favor extractive models, assigning the highest scores to their outputs. The indicators of the ROUGE series are generally accepted by the summarization community for both single and multi document summarization models for various summarization approach. While the other evaluation indicators are just for assistance currently.
	\begin{description}
		\item[ROUGE:] \cite{Lin:2004ROUGE},  (Recall-Oriented Understudy for Gisting Evaluation) evaluates the overlap of textual units (such as n-grams or word sequences) between the generated summary and a set of reference summaries (human-written summaries of the source document).
	 
		\item[S3:] \cite{Peyrard:2017Learning}, introduces a model-based metric that utilizes previously established evaluation metrics, such as ROUGE, JS-divergence, and ROUGE-WE, as input features to predict evaluation scores. The model is trained on human judgment datasets from TAC conferences. 
		\item[BertScore:] \cite{Zhang:2020BERTSCORE}, calculates similarity scores by aligning generated and reference summaries at the token level. Token alignments are determined greedily to maximize the cosine similarity between contextualized token embeddings from BERT. 
		\item[MoverScore:] \cite{Zhao:2019MoverScore}, evaluates the semantic distance between a summary and reference text using the Word Mover's Distance \cite{Werner:2019Speeding}, which operates over n-gram embeddings pooled from BERT representations.
		\item[Sentence Mover’s Similarity (SMS):] \cite{Clark:2019Sentence-mover} extends the Word Mover’s Distance (WMD) \cite{Werner:2019Speeding}, a semantic similarity metric based on word embeddings and optimal transport. SMS combines contextual embeddings with WMD for text generation evaluation, treating documents as a collection of sentence embeddings. It also includes a variation that represents documents as both a collection of sentences and a collection of words.
		\item[SummaQA:] \cite{Scialom:2019Answers-unite} uses a BERT-based question-answering model to answer cloze-style questions based on generated summaries. The questions are created by masking named entities in the source documents linked to the evaluated summaries. The metric provides both the F1 overlap score and the QA-model confidence.
		\item[BLANC:] \cite{Vasilyev:2020BLANC}, introduces a reference-less metric that evaluates the performance improvement of a pre-trained language model when it is provided with a document summary while performing language understanding tasks on the source document’s text.
		\item[SUPERT:] \cite{Gao:2020SUPERT}, introduces a reference-less metric, initially designed for multi-document summarization, that assesses the semantic similarity between model outputs and pseudo-reference summaries. These pseudo-references are created by extracting salient sentences from the source documents, utilizing soft token alignment techniques.
		\item[BLEU:] \cite{Papineni:2002BLEU}, is a corpus-level, precision-focused metric that calculates n-gram overlap between a candidate and reference utterance, incorporating a brevity penalty. It is the primary evaluation metric for machine translation.
		\item[CHRF:] \cite{Popovic:2015CHRF} measures character-based n-gram overlap between model outputs and reference documents.
		\item[METEOR:] \cite{Lavie:2007METEOR}, calculates an alignment between candidate and reference sentences by mapping unigrams in the generated summary to corresponding unigrams in the reference, considering stemming, synonyms, and paraphrastic matches. Precision and recall are computed and reported as a harmonic mean.
		\item[CIDEr:] \cite{Vedantam:2015CIDEr}, calculates {1–4}-gram co-occurrences between the candidate and reference texts, down-weighting common n-grams, and computing the cosine similarity between the n-grams of the candidate and reference texts.
		\item[Data Statistics:] \cite{Grusky:2018bNewsroom}, defines three measures to assess the extractiveness of a dataset:
			\begin{itemize}
				\item \textbf{Extractive fragment coverage}: This measure represents the percentage of words in the summary that are directly taken from the source article, indicating the degree to which a summary is derivative of the original text.
				\item \textbf{Density}: This refers to the average length of the extractive fragment that each word in the summary is associated with.
				\item \textbf{Compression ratio}: This is the ratio of words in the article to words in its summary, reflecting the extent of compression.
				\item \textbf{Novelty Score}: The percentage of n-grams in the summary that are not found in the input document.
				\item \textbf{Redundancy Score}: The percentage of n-grams in the summary that repeat.
			\end{itemize} 
	\end{description}

      \textbf{Factual Consistency Check} \label{FactualCC}
	  The automatic evaluation methods mentioned above cannot be used to measure the factual consistency of a generated summary. A factually consistent summary includes only statements that are entailed by the source document. Fact-checking involves verifying facts against the entire body of available knowledge and identifying discrepancies between the source documents and the generated summary. Errors made by summarization models typically involve the use of incorrect entity names, numbers, and pronouns. Other errors, such as those related to negations and common sense, occur less frequently \cite{Kryscinski:2020Factual}.
Factual inconsistency occur either at the entity level or at the relation level \cite{Nan:2021Entity-level}. 
\begin{itemize}
	\item \textbf{Entity hallucination problem}:
	At the entity level, a generated summary may include named entities that were never present in the source document. For instance, the term "UK" was not mentioned in a source document from the XSUM dataset \cite{Narayan:2018Ranking-sentences}. Though the source document referred to a study involving people from Italy and the Netherlands, the generated summary stated: "People in Italy and the Netherlands are more likely to consume fewer cups of coffee than those in the UK." Here, "UK" in the summary resulted from a model hallucination. 
  \item \textbf{Entity relations identification problem}:
  This type of inconsistency is more difficult to identify. It arises when entities are present in the source document, but the relationships between them are not explicitly stated. For example, summary sentences may paraphrase multiple fragments from the source document, while the source document may use specific linguistic constructs, such as co-reference, to link different parts of the document together. This can lead to inconsistencies in how the entities are related in the summary compared to the source.
\end{itemize}

\cite{Fractalego}, the fact checking implementation aim to predict (TRUE/FALSE) whether a claim is consistent with the provided evidence, it is a GPT-2 \cite{Ziegler:2020Fine-Tuning} model trained on FEVER (dataset for fact verification against textual sources) dataset restricted to the SUPPORTING and REFUTING options. The score gives indications of what probability a summary should be considered factually consistent.
For example, given the following evidence and claim: \\
\textbf{Evidence}: Jane writes code for Huggingface.\\
\textbf{Claim}: Jane is an engineer.\\
Output: Y: 0.95, N: 0.05

\section{Datasets}
Data are crucial for training, testing, and validating machine learning models. However, acquiring training data is a time-consuming and resource-intensive process. Consequently, existing datasets for text summarization tasks are typically restricted to a limited range of domains. The increasing development of large datasets has made it possible for the continuous improvement of text summarization model. However, training is often dominated by bias present in summarization datasets preventing models from learning accurate content selection \cite{Fabbri:2021SummEval}:  
\begin{itemize}
    \item Datasets used for training text summarization models may contain strong layout biases. For instance, in news-related summarization datasets, the structure of news articles often places the most important information in the first paragraph or the first sentence of each paragraph. This structural pattern can introduce positional and extractive biases during training, as the model may learn to prioritize content from these specific positions, potentially overlooking other relevant information in the document.
    \item Datasets like CNN/DailyMail, where each news article is paired with a single reference summary, result in an underconstrained summarization task. Training for summarization in such cases relies on only one human-written summary for each document, which may limit the diversity of summary styles and perspectives. This can lead to models being overly dependent on a specific way of summarizing the content, potentially reducing the generalization ability to create diverse and informative summaries.
    \item Issues associated with automatically collected news datasets also include noisy, low-quality data. This is often due to extraneous information, such as hyperlinks, advertisements, or click-bait descriptions of other articles, which can be present in reference summaries. Such irrelevant content can degrade the quality of the training data, leading to poor model performance and inaccurate or misleading summaries.
	\item The way in which many reference summaries (for example in the CNN/DailyMail dataset) were constructed, by naively concatenating bullet-point summaries into contiguous sequences, which affects the consistency, coherence, and relevance of the reference summary. 
\end{itemize} 
Pre-training on a large text corpora like C4 “Colossal Clean Crawled Corpus” (C4), introduced in \cite{Raffel:2023Exploring} and HugeNew for single document summarization, and NewSHead \cite{Gu:2020News-Stories} for multi document summarization provides the model with a general-knowledge, thus helping the training to transfer learning more effectively to downstream tasks especially when the pretrained data and downstream data domains are aligned.\\
\begin{description}
\item [Dataset for Single Document Summarization:] The following are some dataset for Single Document Summarization task, these datasets contain a set of source documents and a set of reference summary for each document; 
    \begin{itemize} 
	\item XSum \cite{Narayan:2018Topic-aware}, consists of 227k BBC articles from 2010 to 2017 covering a wide variety of subjects along with professionally written single-sentence summaries. 
	\item CNN/DailyMail \cite{Hermann:2015Teaching-machines}, dataset contains 93k articles from the CNN, and 220k articles the Daily Mail newspapers. Both publishers supplement their articles with bullet point summaries. the non-anonymized variant is also available in \cite{See:2017Pointer-Generator}. 
	\item NEWSROOM \cite{Grusky:2018Newsroom}, is a large dataset containing 1.3M article-summary pairs written by authors and editors in the newsrooms of 38 major publications between 1998 and 2017. 
	\item Gigaword \cite{Rush:2015Neural-attention}, contains 4M examples extracted from news articles (seven publishers) from the Gigaword corpus \cite{Graff:2003Gigaword}. 
	\item arXiv, PubMed \cite{Cohan:2018Discourse-aware}, are two long document datasets of scientific publications from arXiv.org (113k) and PubMed (215k). 
	\item BigPatent \cite{Sharma:2019BIGPATENT}, consists of 1.3 million U.S. patents along with human summaries under nine patent classification categories. 
	\item WikiHow \cite{Koupaee:2018Wikihow}, is a large-scale dataset of instructions from the online WikiHow.com website. Each of 200k examples consists of multiple instruction-step paragraphs along with a summarizing sentence. 
	\item Reddit TIFU \cite{Kim:2019Reddit}, contains 120K posts of informal stories from the online discussion forum Reddit, more specifically the TIFU sub-reddit from 2013-Jan to 2018-Mar. The sub-reddit posts strictly follow the rule of writing a descriptive TL;DR summary. 
	\item AESLC \cite{Zhang:2019Email}, consists of 18k email bodies and their subjects from the Enron corpus \cite{Klimt:2004Enron}, which is a collection of email messages exchanged by employees of the Enron Corporation.
	\item BillSum \cite{Kornilova:2019BillSum}, contains 23k US Congressional bills and human-written reference summaries from the 103rd-115th (1993-2018) sessions of Congress.	
	\item BookSum \cite{Kryscinski:2021BookSum}, contains summaries for 142,753 paragraphs, 12,293 chapters and 436 books for long-form narrative summarization. This dataset covers source documents from the literature domain, such as novels, plays and stories, and includes highly abstractive, human written summaries. 
	\item GovReport \cite{Huang:2021Efficient}, it contains long documents (9.4k words) and summaries (553 words), consisting of about 19.5k U.S. government reports with expert-written abstractive summaries.
	\item Multi-LexSum \cite{Shen:2022Muliti-LexSum}, is collection of long document of 9,280 legal case with expert written summaries, each case may exceed two hundred pages.
\end{itemize} 
    CNN/DailyMail, Multi-News, arXiv, PubMed, BigPatent datasets contain input documents length longer than the input length ($L_{input} = 512$ tokens) in Bert. This would present a problem for position embeddings which would never be updated for longer input lengths. Tranformer-based model such as $PEGASUS_{LARGE}$ and BART \cite{Lewis:2019BART}, can generalize well when finetuning up to $L_{input} = 1024$ tokens. Long document summarization models such as Longformer \cite{Beltagy:2020Longformer} and LongT5 \cite{Guo:2022LongT5}, generalize well for BigPatent, arXiv, PubMed, GovReport, Multi-LexSum and Multi-News, whose average input length are well beyond 1024 tokens.

\item [Dataset for Multi Document Summarization:] The primary areas covered by multi-document summarization datasets include news, scientific papers, and Wikipedia. In the early stages of multi-document summarization research, the datasets were relatively small, making them less suitable for training deep neural network models. However, recent datasets are significantly larger and better suited for fine-tuning summarization tasks. These datasets typically consist of a set of source documents paired with corresponding reference summaries.  
 \begin{itemize} 
  
    \item $MS^2$ (Multi-Document Summarization of Medical Studies) \cite{DeYoung:2021MS2}, contain over 470k documents and 20k summaries derived from the scientific literature. This dataset facilitates the development of systems that can assess and aggregate contradictory evidence across multiple studies, and is the first large-scale, publicly available multi-document summarization dataset in the biomedical domain. 
    \item Multi-XScience \cite{Lu:2020Multi-XScience}, is a dataset designed for abstractive multi-document summarization. The source data for Multi-XScience comes from Arxiv and Microsoft Academic Graphs. This dataset exhibits fewer positional and extractive biases compared to the WikiSum and Multi-News datasets, thus partially mitigating the drawback of models scoring higher by copying sentences from specific positions. As a result, Multi-XScience is more suitable for training models that can generate diverse and meaningful summaries.
    \item Multi-News \cite{Fabbri:2019Multi-News}, is a relatively large-scale dataset consisting of news articles and their corresponding human-written summaries, sourced from the site newser.com. The dataset includes 56,216 article-summary pairs and provides trace-back links to the original documents, allowing for a detailed analysis of the content and its summarized version.
    \item WCEP \cite{Gholipour:2020Large-Scale}, the Wikipedia Current Events Portal dataset, contains human-written summaries of recent news events. To extend the input data for large-scale news article generation, similar articles are sourced from the Common Crawl News dataset. The WCEP dataset is well-aligned with real-world industrial use cases, making it suitable for training and evaluating models in practical, large-scale summarization tasks.
    \item Rotten Tomatoes \cite{Wang:2016Network-Based}, is a dataset consisting of reviews for 3,731 movies collected from the Rotten Tomatoes website. The reviews include both professional critics' assessments and user comments. For each movie, a one-sentence summary is provided by professional editors, offering a concise overview of the film's reception. This dataset is commonly used for tasks like sentiment analysis and text summarization.
    \item WikiSum \cite{Liu:2018Generating}, WikiSum is a dataset based on English Wikipedia and targets abstractive multi document summarization. In each instance, the input is comprised of a Wikipedia topic (title of article) and a collection of Golden summaries which are the real Wikipedia articles. The dataset is restricted to the articles with at least one crawlable citation. The official split divides the articles roughly into 80/10/10 for train/development/test subsets, resulting in 1865750, 233252, and 232998 examples respectively.
    \item DUC 2004 \cite{DUC2004},The DUC2004 dataset is a dataset for document summarization. Is designed and used for testing only. It consists of 500 news articles, each paired with four human written summaries. Specifically it consists of 50 clusters of Text REtrieval Conference (TREC) documents, from the following collections: AP newswire, 1998-2000; New York Times newswire, 1998-2000; Xinhua News Agency (English version), 1996-2000. Each cluster contained on average 10 documents.
    \item OpoSum \cite{Angelidis:2018Opinions}, is a dataset that collects multiple reviews from six product domains on Amazon. In addition to containing reviews and corresponding summaries, the dataset also includes information about each product's domain and polarity, which can serve as auxiliary supervision signals. A subset of the dataset has been manually annotated; for each domain, 10 different products were sampled (across various ratings), with 10 reviews per product, totaling 600 reviews. These reviews are divided into development (300) and testing (300) sets.
    \item Healthline \cite{Shah:2021Nutri-bullets}, Healthline is a nutrition related dataset for multi-document summarization, using scientific studies.
    \item Aquamuse \cite{Kulkarni:2020AQuaMuSe},  consists of 5,519 query-based summaries, each associated with an average of 6 input documents. These input documents are selected from an index of 355 million documents sourced from Common Crawl. The dataset is designed for tasks like query-based summarization, where the goal is to generate concise summaries based on specific queries related to large collections of documents.
    \item GameWikiSum \cite{Antognini:2020GameWikiSum}, GameWikiSum is a domain-specific (video game) dataset for multi-document summarization, which is one hundred times larger than commonly used datasets, and in another domain than news. Input documents consist of long professional video game reviews as well as references of their gameplay sections in Wikipedia pages.
    \item LSARS (Large Scale Abstractive multi-Review Summarization) \cite{Pan:2020LSARS}, is large-scale abstractive multi-review summarization dataset that leverages more than 17.9 billion raw reviews and uses novel aspect-alignment techniques based on aspect annotations. 
    \item Opinosis \cite{Ganesan:2010Opinosis}, is a dataset that contains reviews from 51 topic clusters collected from websites like TripAdvisor, Amazon, and Edmunds. Each topic cluster includes approximately 100 sentences on average, with reviews gathered from various sources. For each cluster, five professionally written gold-standard summaries are provided, which are used for model training and evaluation. This dataset is commonly used for abstractive summarization tasks, particularly in the context of review summarization.
      \end{itemize}

\end{description}

\section{Experiment and Results} \label{Experiment}
 In this section we show the test case of abstractive text summarization approach, by comparing the recent abstractive text summarization models on some datasets, verifying the claims about document domain bias in text summarization that is, the effect of training a model on dataset of certain document domain with respect to the domain of the documents in the test case. In addition we compare the factual consistency level of the summaries generated in other to verify the level of factual inconsistency as claimed by \cite{Cao:2018Fact-Aware}; \cite{Goodrich:2019Factual-Accuracy}; \cite{Falke:2019Ranking-Generated}; \cite{Kryscinski:2019Neural-Text}.
 The experiment consiste of pre-trained transformer-based model for text summarization task, pre-trained on large corpus and then finetuned on data pairs (Document-summary).  \textbf{BART}, \textbf{PEGASUS}, \textbf{Longformer}, \textbf{LongT5}, \textbf{REFLECT}, \textbf{PRIMERA}, and \textbf{CENTRUM}, are the recently pubblished transformer based abstractive text summarization models selected for experiment. We report the results of the evaluation of each model using automatic metrics like \textbf{ROUGE}, \textbf{METEOR}, \textbf{CHRF}, and \textbf{BertScore}.
 The factual consistency check was computed with \cite{Fractalego} for each generated summary (claim) using the source document as evidence. The use of GPT-2 \cite{Ziegler:2020Fine-Tuning} in \cite{Fractalego}'s factual consistency check implementation constrains the maximum word length of the evidence to 1024 and the claim to 420 words. To effectuate the factual consistency check for long sequence/document, and multi documents, when possibile the reference summary was used as evidence, in this case it is assumed that the reference of the source document is a factual consistent summary in terms of entities and the relation between entities. When the reference summary is longer than the required sequence length, the factual consistency check was not computed for the generated summary. The score obtained with this assumption serves as a possible approximation of factual consistency check for long document and multi document summarization models.

 Experiments and evaluations using open-source implementation of automatic evaluation via Hugging-Face \cite{HuggingFace}, publicly available trained weights on Hugging-Face \cite{Wolf:2020Transformers} and \cite{GitHub}, and publicly available \cite{Fractalego}'s factual consistency check weights on Hugging-Face\footnote{Hugging Face is a community and data science platform that provides: Tools that enable users to build, train and deploy ML models based on open source (OS) code and technologies}.
 Table \ref{table-exper} shows the breakdown of the experimented models and the datasets on which they were finetuned on which are divide into single document and multi document summarization.
\begin{table}[ht]
	\centering
	\label{table-exper}
	\caption{The breakdown of the experimented finetuned models.}
		\vspace*{2mm}	

	\resizebox{0.99\linewidth}{!}{
	\begin{tabular}{|c|c|c|c|c|c|c|c|}				
		
\hline
\textbf{Dataset}&\multicolumn{4}{|c|}{\textbf{Single document}}& \multicolumn{3}{|c|}{\textbf{Multi document}}\\
\hline	
&\multicolumn{2}{|c|}{\textbf{Short document}} & \multicolumn{2}{|c|}{\textbf{Long document}}& & & \\
\hline
&\textbf{BART}	&	\textbf{PEGASUS}  &  \textbf{Longformer}	 & \textbf{LongT5}  & \textbf{CENTRUM} & \textbf{PRIMERA} &	\textbf{REFLECT}  \\
\hline
\textbf{xsum}   & \checkmark  & \checkmark &  &  &  & &  \\
\hline
\textbf{cnn-dailymail}  & \checkmark  & \checkmark  & \checkmark &  &  & &  \\
\hline
\textbf{newsroom} &   &\checkmark  &  &  &  & &  \\
\hline
\textbf{multi-news}  & \checkmark  & \checkmark &  &  & \checkmark & \checkmark & \checkmark \\
\hline
\textbf{gigaword}  &   &\checkmark  &  &  &  & &  \\
\hline
\textbf{wikihow}  &   & \checkmark &  &  &  & &  \\
\hline
\textbf{reddit tifu}  &   &\checkmark  &  &  &  & &  \\
	\hline
	\textbf{big patent}   &   & \checkmark &  &  &  & &  \\
\hline	
\textbf{arxiv} &   & \checkmark & \checkmark &  &  & &  \\
\hline
\textbf{pubmed}  &   & \checkmark & \checkmark & \checkmark &  & &  \\
\hline
\textbf{aeslc}  &   & \checkmark  &  &  &  & &  \\
\hline
\textbf{booksum} &   &  &  \checkmark& \checkmark &  & &  \\
\hline
\textbf{amazonreviews}  & \checkmark  &  &  &  &  & &  \\
\hline

\end{tabular}	
}
\end{table}

\subsection{Single document summarization results}
  
\textbf{Short document summarization results:}
We use finetuned weights publicly available on Hugging-Face hub. We denoted each fine-tuned model for short document summarization with $Ms_x$. We run 7 short document samples from various document domain on each finetuned model.
 \begin{itemize}
 
				\item \textbf{$Ms_1$ - BART large}: \textit{facebook/BART\mbox{-}large\mbox{-}cnn}, BART large fine-tuned on \textit{cnn} dataset.
				\item \textbf{$Ms_2$ - BART large}: \textit{sumedh/distilBART\mbox{-}cnn\mbox{-}12\mbox{-}6\mbox{-}amazonreviews}, distilled BART large (Bart large pretrained weight fine-tuned with cnn dataset for a 12 layer encoder and 6 layer decoder model) fine-tuned on \textit{amazonreviews} dataset. 
				\item \textbf{$Ms_3$ - BART large}: \textit{facebook/BART\mbox{-}large\mbox{-}xsum}, BART large fine-tuned on \textit{xsum} dataset.
				\item \textbf{$Ms_4$ - PEGASUS large}: \textit{$google/PEGASUS\mbox{-}large\mbox{-}cnn\_dailymail$}, PEGASUS large fine-tuned on $cnn\_dailymail$ dataset. 
				\item \textbf{$Ms_5$ - PEGASUS large}: \textit{google/PEGASUS\mbox{-}large\mbox{-}xsum}, PEGASUS large fine-tuned on \textit{xsum} dataset. 
				\item \textbf{$Ms_6$ - PEGASUS large}: \textit{google/PEGASUS\mbox{-}large}, PEGASUS large fine-tuned on \textit{xsum, $cnn\_dailymail$, newsroom, $multi\_news$, gigaword, wikihow, $reddit\_tifu, big\_patent$, arxiv, pubmed, aeslc} datasets. 
				\item \textbf{$Ms_7$ - PEGASUS large}: \textit{google/PEGASUS\mbox{-}large\mbox{-}xsum}, PEGASUS large fine-tuned on xsum dataset. 
				\item \textbf{$Ms_8$ - BART large}: \textit{ $datien228/distilBART\mbox{-}cnn\mbox{-}12\mbox{-}6\mbox{-}ftn\mbox{-}multi\_news$}; \\ sshleifer/distilbart\mbox{-}cnn\mbox{-}12\mbox{-}6 (Bart large pretrained weight fine-tuned with cnn dataset for a 12 layer encoder and 6 layer decoder model) fine-tuned on the English portion of multi-news dataset. 
 \end{itemize}
 \begin{table}[ht]
	\centering
	\caption{The average score results from seven short document samples, spanning various domains, generated by fine-tuned models for abstractive short-document summarization.}
	\vspace*{2mm}
	\resizebox{0.99\linewidth}{!}{
	\begin{tabular}{|c|c|c|c|c|c|c|c|c|c|}
									
									\hline
							
							\textbf{Metric}	& $\mathbf{Ms_1}$ &\textbf{$\mathbf{Ms_2}$ }	& \textbf{$\mathbf{Ms_3}$ } & \textbf{$\mathbf{Ms_4}$ }  &	\textbf{$\mathbf{Ms_5}$ }	 & \textbf{ $\mathbf{Ms_6}$}  & \textbf{ $\mathbf{Ms_7}$}	 &\textbf{$\mathbf{Ms_8}$ } \\
						\hline
						\textbf{ROUGE-1}	&	0.2292 & 0.233&  0.1117&  0.2368&  0.0534&  0.2896&  0.1236&  0.32\\
							\hline
							\textbf{ROUGE-2}  &	0.05&  0.076&  0.0261&  0.0652&  0.0135& 0.0804&  0.0111&  0.1175\\
							\hline
							\textbf{ROUGE-3}  &	0.0205&  0.0477&  0.0105&  0.0286&  0.0072&  0.0426&  0.0023&  0.0728\\
							\hline
							\textbf{ROUGE-L} 	& 	0.1318&  0.1489&  0.0835&  0.1524& 0.0461&  0.1836&  0.0786&  0.1925\\
							\hline
							\textbf{CHRF-1} &		0.2348&  0.209&  0.0961&  0.219&  0.0679&  0.3153&  0.0913&  0.3463\\
							\hline
							\textbf{METEOR}  &	0.1511&  0.1823&  0.0645&  0.1707&  0.0186&  0.2131&  0.0565&  0.2141\\
							\hline
							\textbf{BertScore}  &	0.8369&  0.8349&  0.8364&  0.8372&  0.8004&  0.8493&  0.8267&  0.8372\\
							\hline
							\textbf{FactCheck}  &	0.95& 1.0& 0.935& 0.95& 0.96& 0.985& 0.975& 0.93\\

								\hline
							
				\end{tabular}
						
				}

			\label{evaluate_1}
		\end{table}		

	Table \ref{evaluate_1} show the average score results from seven short document samples, spanning various domains, generated by fine-tuned models for abstractive short-document summarization. Evaluated using ROUGE-1, ROUGE-2, ROUGE-3, ROUGE-L, CHRF-1, METEOR, BertScore automatic metrics, and factual consistency check on the summaries generated by the finetuned short document summarization models.
	
  The evaluation and output of each finetuned model on each of the seven short document sampled is availability in the supplementary \cite{GitRepository}. The results show that model parameter size such as encoder and decoder layer size influence the summary; large models tend to perform better. 
	 Large models fine-tuned using distilation knowledge \cite{Niccolai:2020Distillation} such as $Ms_2$ and $Ms_8$, which are finetuned on dataset with document domain different from the one in which the Teacher model \cite{Niccolai:2020Distillation} was finetuned, proved to be a very valid improvement to abstractive text summarization model. 
	 Models such as $Ms_6$, $Ms_2$ and $Ms_8$ finetuned on multiple datasets which contains documents that cut across various domain showed significant score.
	   The generated summary showed that the repetition problem faced by abstractive model is at its minimum with BART which uses a no\_repeat\_n-gram\_size = 3; a word can only be repeated for a maximum of 3 consecutive times within a sentence.
	   The result of the factual consistency checker show that the factuality problem faced by abstractive text summarization models has reduced significantly compared to the $\sim30\%$ factual inconsistency reported by \cite{Cao:2018Fact-Aware}; \cite{Goodrich:2019Factual-Accuracy}; \cite{Falke:2019Ranking-Generated}; \cite{Kryscinski:2019Neural-Text}. $Ms_2$ shows 100\% factually consistent summaries score.\\

\textbf{Long document summarization results:}
We use finetuned weights publicly available on Hugging-Face hub. We denoted each fine-tuned model for long document summarization with $Ml_x$. We run 2 long document samples from various document domain on each finetuned model.

	\begin{itemize}
			
				\item \textbf{$Ml_1$ - LED\mbox{-}large}: allenai/led\mbox{-}large\mbox{-}16384\mbox{-}arxiv, Longformer large fine-tuned on the arXiv dataset.
				\item \textbf{$Ml_2$ - LED\mbox{-}base}: pszemraj/led\mbox{-}base\mbox{-}16384\mbox{-}finetuned\mbox{-}booksum, Longformer base fine-tuned on the booksum dataset. 
				\item \textbf{$Ml_3$ - LED\mbox{-}large}: patrickvonplaten/led\mbox{-}large\mbox{-}16384\mbox{-}pubmed, Longformer large fine-tuned on the pubmed dataset. 
				\item \textbf{$Ml_4$ - distil\mbox{-}LED\mbox{-}large}: HHousen/distil\mbox{-}led\mbox{-}large\mbox{-}cnn\mbox{-}16384, Longformer initialized from sshleifer/distilBART\mbox{-}cnn\mbox{-}12\mbox{-}6; BART large pretrained weight fine-tuned with cnn dataset for a 12 layer encoder and 6 layer decoder model. Position embedding matrix was copied 16 times in other to process 16K tokens.
				\item \textbf{$Ml_5$ - LED\mbox{-}large}: pszemraj/led\mbox{-}large\mbox{-}book\mbox{-}summary, Longformer encoder and decoder large fine-tuned on the booksum dataset. 
				\item \textbf{$Ml_6$ - LongT5\mbox{-}base}: pszemraj/long\mbox{-}t5\mbox{-}tglobal\mbox{-}base\mbox{-}16384\mbox{-}book\mbox{-}summary, LongT5 base fine-tuned on the booksum dataset. 
				\item \textbf{$Ml_7$ - LongT5\mbox{-}large}: Blaise\mbox{-}g/longt5\mbox{-}tglobal\mbox{-}large\mbox{-}sumpubmed, LongT5 large fine-tuned on sumpubmed dataset using weights from $Ml_8$. 
				\item \textbf{$Ml_8$ - LongT5\mbox{-}large}: Stancld/longt5\mbox{-}tglobal\mbox{-}large\mbox{-}16384\mbox{-}pubmed\mbox{-}3k\_steps, LongT5 large fine-tuned on the pubmed dataset.

\end{itemize}

					\begin{table}[ht]
							\centering
							\caption{The average score results from two long document samples, spanning various domains, generated by fine-tuned models for abstractive long-document summarization.}

								\vspace*{2mm}
						\resizebox{0.99\linewidth}{!}{
							\begin{tabular}{|c|c|c|c|c|c|c|c|c|}
								
								\hline
							
							\textbf{Metric}	& $\mathbf{Ml_1}$ & $\mathbf{Ml_2}$ 	& \textbf{$\mathbf{Ml_3}$ } & \textbf{$\mathbf{Ml_4}$ }  &	\textbf{$\mathbf{Ml_5}$ }	 & \textbf{ $\mathbf{Ml_6}$}  & \textbf{ $\mathbf{Ml_7}$} & \textbf{ $\mathbf{Ml_8}$}	 \\
						\hline
						\textbf{ROUGE-1}	&	0.3468 & 0.3432 & 0.4273 & 0.2735 & 0.3368 & 0.3776 & 0.3766 & 0.1452\\
							\hline
							\textbf{ROUGE-2}  &	0.145& 0.0885& 0.1934 & 0.0775 & 0.1089& 0.1237 & 0.1451 & 0.0655 \\
							\hline
							\textbf{ROUGE-3}  &		0.0936 & 0.0303 & 0.1189& 0.0228 & 0.0472 & 0.0442 & 0.0799 & 0.0392 \\
							\hline
							\textbf{ROUGE-L} 	& 0.2147 & 0.1842 & 0.2515 & 0.1812 & 0.1776 & 0.2158 & 0.2195 & 0.1164\\
							\hline
							\textbf{CHRF-1} &	0.3475 & 0.2983 & 0.3922& 0.1886& 0.3194 & 0.3817 & 0.3517 & 0.0734 \\
							\hline
							\textbf{METEOR}  &	0.2593 & 0.2178 & 0.3019 & 0.1329 & 0.2584 & 0.2181 & 0.2509 &  0.0618\\
							\hline
							\textbf{BertScore}  &	0.8232 & 0.8326 & 0.8498 & 0.8374 & 0.8411 & 0.8492 & 0.8399 & 0.8292\\
							\hline
							\textbf{FactCheck}  &0.97& 0.9455& 0.965& 0.9636& 0.795& 0.9722& 0.8591&  0.9864\\
								\hline
							
							\end{tabular}
							}

							\label{evaluatel1}
						\end{table}		
						
						Table \ref{evaluatel1} show the average score results from two long document samples, spanning various domains, generated by fine-tuned models for abstractive long-document summarization. Evaluated using ROUGE-1, ROUGE-2, ROUGE-3, ROUGE-L, CHRF-1, METEOR, BertScore automatic metric, and factual consistency check on the summaries generated by the finetuned long document summarization models.
            The evaluation and output of each finetuned model on each of the two long document sampled is availability in the supplementary \cite{GitRepository}. The result shows that $Ml_3$ outperforms other finetuned model showing that Longformer \cite{Beltagy:2020Longformer} when finetuned on large dataset like pubmed and on various document domain prove to improve performance of abstractive text summarization for long documents.
						$Ml_4$ shows interesting results though it adapted multiple BART finetuned on cnn dataset such that it is able to process long sequence taking a clue from Longformer.
	   The generated summary showed that the repetition problem faced by abstractive model is at its minimum with Longformer which uses a no\_repeat\_n-gram\_size = 3 from BART.
	  There is need for a more adequate factual consistency check for long document, the result of the factual consistency checker using the reference summary as evidence and the generated summary as claim show that the factuality problem faced by abstractive text summarization models has reduced significantly, $Ml_5$ though showing the least factual consistency score it show some improvement compared to the $\sim30\%$ factual inconsistency reported by \cite{Cao:2018Fact-Aware}; \cite{Goodrich:2019Factual-Accuracy}; \cite{Falke:2019Ranking-Generated}; \cite{Kryscinski:2019Neural-Text}.

\subsection{Multi document summarization results}

We use finetuned weights publicly available on Hugging-Face hub, while REFLECT from github. We denoted each fine-tuned model for multi document summarization with $Mm_x$. We run 10 short document samples from news domain on each finetuned model. Each single document within the multi document from multi-news dataset is separated by the characters $\mid\mid\mid\mid\mid$.

\begin{itemize}
			
				\item \textbf{$Mm_1$ - PRIMERA}: allenai/PRIMERA\mbox{-}multinews, PRIMERA large fine-tuned on the multi\mbox{-}news dataset.
				\item \textbf{$Mm_2$ - CENTRUM}: ratishsp/Centrum\mbox{-}multinews, CENTRUM fine-tuned on the multi\mbox{-}news dataset.
				\item \textbf{$Mm_3$ - REFLECT}: yunzhusong/NAACL2022\mbox{-}REFLECT, REFLECT fine-tuned on the multi\mbox{-}news dataset.
\end{itemize}

		\begin{table}[h]
							\centering
							
							\caption{The average score results from ten multi document samples of news document domain, generated by fine-tuned models for abstractive multi-document summarization.}
							\vspace*{2mm}
							\begin{tabular}{|c|c|c|c|}
								
									\hline

								\textbf{Metric}	& $\mathbf{Mm_1}$ & $\mathbf{Mm_2}$ 	& \textbf{$\mathbf{Mm_3}$ }	 \\
							\hline
							\textbf{ROUGE-1}	&	0.3856& 0.4654& 0.2713  \\
								\hline
								\textbf{ROUGE-2}  &	0.1373&0.1795&0.0315   \\
								\hline
								\textbf{ROUGE-3}  &		0.0848&0.1102& 0.0069    \\
								\hline
								\textbf{ROUGE-L} 	& 	0.19&0.2482& 0.1286\\
								\hline
								\textbf{CHRF-1} &		0.3488& 0.4222& 0.2555   \\
								\hline
								\textbf{METEOR}  &	0.2821& 0.3296& 0.1972  \\
								\hline
								\textbf{BertScore}  &	0.8434& 0.864& 0.7743  \\
								\hline
								\textbf{FactCheck}  &		0.8938 &   0.93  &    0.835 \\
									\hline
							\end{tabular}
							
							\label{evaluatem1}
						\end{table}

	Table \ref{evaluatem1} show the average score results from ten multi document samples of news document domain, generated by fine-tuned models for abstractive multi-document summarization. Evaluated using ROUGE-1, ROUGE-2, ROUGE-3, ROUGE-L, CHRF-1, METEOR, BertScore automatic metric, and factual consistency check on the summaries generated by the finetuned multi document summarization models.

  The evaluation and output of each finetuned model on two of ten multi document sampled is availability in the supplementary \cite{GitRepository}. The results show that $Ml_2$ outperforms other finetuned model and show that the objective function of using the centroid as synthetic summary from the cluster of aggregated salient information in a multi document, brings important improvement to the multi document summarization.
			 The generated summary showed that the repetition problem faced by abstractive text summarization model is at its minimum with PRIMERA ($Mm_1$)and CENTRUM ($Mm_2$) which uses no\_repeat\_n-gram\_size = 3 from BART.
				Extractive-then-Abstractive approach using hierarchical graph used by REFLECT ($Mm_3$) generated not very fluent summaries. 
				There is need for a more adequate factual consistency check for multi document, the result of the factual consistency checker using the reference summary as evidence and the generated summary as claim show that the factuality problem faced by abstractive text summarization models has reduced significantly compared to the $\sim30\%$ factual inconsistency reported by \cite{Cao:2018Fact-Aware}; \cite{Goodrich:2019Factual-Accuracy}; \cite{Falke:2019Ranking-Generated}; \cite{Kryscinski:2019Neural-Text}. 
		
\section{Open challenges of abstractive text summarization}	

Factual faithfulness remains an ongoing research challenge in abstractive text summarization models. A critical challenge in deploying abstractive text summarization models for real-world applications is that they may generate factually inconsistent content relative to the input or even produce entirely "hallucinated" information \cite{Lee:2018Hallucinations}; \cite{Maynez:2020Faithfulness-Factuality}; \cite{Rohrbach:2018Object}; \cite{Zhao:2020Hallucinations}. Though factual inconsistency has been reduced significantly, the available evaluation metric, and factual consistency checkers are not able to wholesomely study the factual faithfulness problem especially for long document and multi document. 
    The document length (long document) that the state of the art abstractive text summarization models can process is still very much limited, real world application may require a model that can summarize a versatile length of input sequence like several books each consisting of hundreds of pages. 
    Large data are only available for limited domains, though more large data have been made available in recent years, there is still need for large data from a variety of document domain, and for a variety of language, both for pre-training and finetuning. 
     Training and finetuning text summarization models is time consuming and resource-intensive specially for model with very large parameters. Distilled knowledge can reduced cost in terms of speed-up and inference time. 
     Human evaluators still remains the reference in evaluating text summarization model though there have been several research in this area, leaving ROUGE as the default automatic metric for text summarization and the other metric as assistant currently.

\section{Conclusions}
In a world where millions of articles are published every day, which can be overwhelming for readers/researchers to follow, the goal of getting computers to generate a summary is still a mission in progress, and some advancement has been recorded with the continuous improvement in deep learning models as well as the increasing availability of large data, with state of the art models benefiting from relatively large amounts of unlabeled training data for pre-training, and leverage transfer learning during finetuning. 
The experiment conducted in this study showed the improvement that pre-trained model has contributed to abstractive text summarization task by evaluating how these model's performance may rely intrinsically on the difference in input document length, dataset domain, parameters, and pre-training objectives.
We showed the performance of state of the art pre-trained transformer-based models such as BART and PEGASUS for short documents summarization, Longformer and LongT5 for long documents summarization, CENTRUM and PRIMERA, REFLECT for multi document summarization, and obtained interesting results enough to encourage more research in abstractive text summarization.


\section{Data Availability}
The experiments, source codes, model parameters and the supplementary results are available in \cite{GitRepository}, The data in the test case is drawn from Multi-news, Arxiv, Pubmed, CNN, WikiHow, Big-Patent and Bill-Sum datasets available in \cite{HuggingFace}.

\bibliographystyle{unsrt}  
\bibliography{references2}

\end{document}